\documentclass{article}

\usepackage[final, nonatbib]{neurips_2021_ml4ad}

\usepackage[utf8]{inputenc} % allow utf-8 input
\usepackage[T1]{fontenc}    % use 8-bit T1 fonts
\usepackage{hyperref}       % hyperlinks
\usepackage{url}            % simple URL typesetting
\usepackage{booktabs}       % professional-quality tables
\usepackage{amsfonts}       % blackboard math symbols
\usepackage{nicefrac}       % compact symbols for 1/2, etc.
\usepackage{microtype}      % microtypography
\usepackage{xcolor}         % colors

\usepackage[utf8]{inputenc} % allow utf-8 input
\usepackage[T1]{fontenc}    % use 8-bit T1 fonts
\usepackage{url}            % simple URL typesetting
\usepackage{booktabs}       % professional-quality tables
\usepackage{amsfonts}       % blackboard math symbols
\usepackage{nicefrac}       % compact symbols for 1/2, etc.
\usepackage{microtype}      % microtypography
\usepackage{pifont}

\usepackage{tabularx}
\newcolumntype{L}{>{\raggedright\arraybackslash}X}
\usepackage{array, makecell} \newcolumntype{M}[1]{>{\centering\arraybackslash}m{#1}}
\usepackage{diagbox}
\usepackage{caption}
\usepackage{float}
\usepackage[export]{adjustbox}
\usepackage{comment}
\usepackage{times}
\usepackage{epsfig}
\usepackage{graphicx}
\usepackage{amsmath}
\usepackage{amssymb}
\usepackage{listings}
\usepackage{xcolor}

\title{PKCAM: Previous Knowledge Channel Attention Module}

\author{%
  Eslam Mohamed Bakr \\
  Deep Learning Researcher\\
  Valeo R\&D Cairo, EGYPT\\
  Cairo University, Egypt \\
  \texttt{eslam.mohamed-abdelrahman@valeo.com} \\
  \And
  Ahmad El Sallab \\
  AI Senior Expert \\
  Valeo R\&D Cairo, EGYPT \\
  \texttt{ahmad.el-sallab@valeo.com} \\
  \And
  Mohsen A. Rashwan \\
  Professor \\
  Electronics and Communications Department, Cairo University, Egypt \\
  \texttt{mrashwan@rdi-eg.com} \\
}

\begin{document}

\maketitle

\begin{abstract}
Recently, attention mechanisms have been explored with ConvNets, both across the spatial and channel dimensions.
However, from our knowledge, all the existing methods devote the attention modules to capture local interactions from a uni-scale.
In this paper, we propose a Previous Knowledge Channel Attention Module(PKCAM), that captures channel-wise relations across different layers to model the global context.
Our proposed module PKCAM is easily integrated into any feed-forward CNN architectures and trained in an end-to-end fashion with a negligible footprint due to its lightweight property.
We validate our novel architecture through extensive experiments on image classification and object detection tasks with different backbones.
Our experiments show consistent improvements in performances against their counterparts. Our code is published at https://github.com/eslambakr/EMCA.
\end{abstract}

% -------------------------------------------------------------------------------
%                                Introduction
% -------------------------------------------------------------------------------
\section{Introduction}

% ConvNets
Over the years, CNN architectures have evolved with many ideas to better deal with spatial image features. Moreover, their localized nature makes such features lack the global view of the image. Deeper architectures emerged that stack multiple convolution layers, known with different names; backbone, bottleneck, feature extractor, or encoder. The main feature of such architectures is their ability to cover spatial features at multiple scales. As we go deeper, the feature maps get smaller, while their content represents a wider region in the space, which gets us closer to better semantics of the image contents \cite{luo2016understanding}.
With the emergence of AlexNet \cite{krizhevsky2012imagenet}, many kinds of research investigate to further improve the performance of deep CNNs. \cite{simonyan2014very} \cite{he2016deep} \cite{szegedy2017inception} \cite{szegedy2016rethinking} \cite{srivastava2015training} have sought to strengthen the CNNs by making it deeper and deeper as they have shown that increasing the depth of a network could significantly increase the quality of the learned representations.
% Transition between CNNs to Attention
Many researchers are continuously investigating to further improve the performance of deep CNNs by incorporating attention mechanisms to exploit its ability to cover the relationship between the learned spatial features.

% Why attention, Feature representation
Attention modules, in general, are designed to suppress noise while keeping useful information by refining the learned features using attention scaling.
By quoting from the human perception process \cite{mnih2014recurrent} where the high-level information is used in guiding the bottom-up learning process by capturing more sophisticated features while disregarding irrelevant details.
Human perception and visual attention \cite{beck2009top} \cite{desimone1998visual} \cite{mnih2014recurrent} \cite{desimone1995neural} is enhanced by top-down stimuli and non-relevant neurons will be suppressed in feedback loops.
Referencing to human visual system, various different attention mechanisms \cite{cao2019gcnet} \cite{wang2018non} \cite{zhao2020exploring} \cite{vaswani2017attention} \cite{wang2017residual} have been explored and integrated into deep CNNs.
Attention mechanisms were introduced in the context of CNNs to capture the relations between features, either across the spatial dimension as in \cite{carion2020end} \cite{hu2018gather} ,or across channel-wise dimension as in  \cite{wang2018non} \cite{cao2019gcnet} \cite{hu2018squeeze} \cite{lee2019srm} \cite{Wang_2020_CVPR} or across both dimensions as in \cite{park2018bam} \cite{woo2018cbam} \cite{fu2019dual} \cite{wang2017residual} \cite{linsley2018learning} \cite{roy2018recalibrating} \cite{chen2017sca}.
% Transition between Attention to PKCAM
Although these attention methods have achieved higher accuracy than their counterpart baselines which do not invoke any attention mechanisms in their architectures, they often bring higher model complexity and exploit only the current feature map while refining it, that's why we call it local attention mechanisms.

% Importance of Global/Previous Knowledge
Exploiting previous knowledge has been applied to image classification \cite{huang2017densely} \cite{iandola2014densenet}, image segmentation \cite{ronneberger2015u}, tracking \cite{ma2015hierarchical}, and human pose estimation \cite{newell2016stacked} where they obtain enhanced performance.
DenseNets \cite{huang2017densely} encourage feature reuse by connecting each layer to every other layer in a feed-forward fashion.
U-Net \cite{ronneberger2015u} consists of two paths, which are contracting path to capture context and a symmetric expanding path that enables precise localization, where feature reuse is introduced through using skip connection between two paths.
Driven by the significance of employing feature reuse while learning different tasks \cite{szegedy2015going} \cite{huang2017densely} \cite{iandola2014densenet} \cite{chen2016attention} \cite{ma2015hierarchical} \cite{newell2016stacked}, a question arises: How can one incorporate previous knowledge aggregation while learning channel attention more efficiently?

% Our work
To answer this question, we introduce PKCAM, a novel feature recalibration module based on channel attention, which improves the quality of the representations produced by a network using the global information to selectively emphasize informative features and suppress less useful ones. In contrast to the aforementioned attention mechanisms, our global context aware attention block obtains additional inputs from all preceding attention blocks, that have the same depth, and passes on its refined feature-maps to all subsequent blocks, creating global awareness from exploiting previous knowledge aggregation from earlier layers that can capture fine-grained information which is useful for precise localization while attending to features from earlier layers that can encode abstract semantic information, which is robust to target appearance changes.

The contributions of this paper are summarized as follows:
\begin{itemize}
    \item  We propose a simple and effective attention module, PKCAM, which can be integrated easily with any CNNs and applied across all it's blocks due to the lightweight computation of our novel architecture.
    \item We verify the effectiveness and robustness of PKCAM throughout extensive experiments with various baseline architectures on KITTI dataset.
    \item Through detailed analysis along with ablation studies, we examine the internal behavior and validity of our method.
\end{itemize}
% -------------------------------------------------------------------------------
%                                Related Work
% -------------------------------------------------------------------------------
\section{Related work}
\label{Related work}

\tabcolsep=0.08cm
\begin{table}[ht]
\fontsize{8}{7.2}\selectfont.
  \caption{Comparison of channel attention module by projecting each channel mechanism according to the abstract global context modeling skeleton that was introduced by GCNet [1] in Figure 4(a).}
  \label{Tab_Equ_channel_att}
  \centering
  \begin{tabular}{llll}
    \toprule
     \hfil Methods &\hfil Context Modeling &\hfil Transform &\hfil Fusion \\
    \midrule
     SE \cite{hu2018squeeze} & $Y_c= GAP(X)$\refstepcounter{equation}(\theequation)\label{SE_C}&  $Z=\sigma(F.C_c(Relu(F.C_{\frac{c}{r}}(Y_c))))$\refstepcounter{equation}(\theequation)\label{SE_T}& $F= Z \odot X $\refstepcounter{equation}(\theequation)\label{SE_F}\\
     ECA \cite{Wang_2020_CVPR} & $Y_c= GAP(X)$\refstepcounter{equation}(\theequation) \label{ECA_C}& $Z= \sigma(C1D_k(Y_c))$\label{ECA_T} &  $F= Z \odot X$ \refstepcounter{equation}(\theequation)\label{ECA_F} \\
     SRM \cite{lee2019srm} & $Y_{c*d}= \text{concat}[GAP(X),SP(X)]$\refstepcounter{equation}(\theequation)\label{SRM_C} & $Z= \sigma(\sum_d(Y_{c*d} \odot W_{c*d}) )$ \refstepcounter{equation}(\theequation)\label{SRM_T}&  $F= Z \odot X$ \refstepcounter{equation}(\theequation)\label{SRM_F}\\
     GC \cite{cao2019gcnet} & $Y_c= X_{C*HW} \otimes S.M. (C1D_k(X))$\refstepcounter{equation}(\theequation)\label{GC_C} & $Z= F.C_c(Relu(F.C_{\frac{c}{r}}(Y_c)))$\refstepcounter{equation}(\theequation)\label{GC_T}& $F= Z + X$ \refstepcounter{equation}(\theequation)\label{GC_F}\\
     %GSoP \cite{gao2019global} & ? & ? & ? \\
    \bottomrule
  \end{tabular}
\end{table}
% The abstract global context modeling framework
The basic channel attention module \cite{hu2018squeeze} \cite{cao2019gcnet} \cite{wang2018non} \cite{lee2019srm} \cite{Wang_2020_CVPR} \cite{gao2019global} aims at strengthening the output features of one convolution block, $\textbf{X} \in \mathbb{R}^{H\times W\times C}$, where W, H and C are width, height and channel dimension.
Table \ref{Tab_Equ_channel_att} projects each channel mechanism according to the abstract global context modeling skeleton that was introduced by GCNet \cite{cao2019gcnet} in Figure 4(a).
The abstract global context modeling framework \cite{cao2019gcnet} consists of three main blocks which are context modeling, transform and fusion blocks.

% Mapping SE and ECA
SE-Net \cite{hu2018squeeze} and ECA-Net \cite{Wang_2020_CVPR} squeeze the spatial dimension of the output features of one convolution block $\textbf{X}$ via aggregating it through channel-wise global average pooling (GAP), where $Y_c = \frac{1}{WH} \sum_{i=1,j=1}^{W,H} X_{ij}$, then SE-Net use non-linear transformation using two fully connected layers with a dimensionality reduction propriety to control the model complexity expansion, the transformation block is called excitation.
ECA-Net \cite{Wang_2020_CVPR} analyzing effects of dimensionality reduction at the SE-Net transformation block that shows that while dimensionality reduction can reduce model complexity, it destroys the direct correspondence between the channel and its weight. Therefore ECA-Net employs a more efficient transformation function as shown in Table \ref{Tab_Equ_channel_att}, where a one-dimensional convolution layer (C1D) is used which only involves k parameters that guarantees both efficiency and effectiveness. 
SE-Net \cite{hu2018squeeze} and ECA-Net \cite{Wang_2020_CVPR} opt to employ a simple gating mechanism with a sigmoid activation function $\sigma$ that produces a channel weights $Z$. The third block of the abstract global context modeling framework \cite{cao2019gcnet} is fusing the channel weights $Z$ with the original feature map $X$ to produce a recalibrated feature map $F$. Both SE-Net \cite{hu2018squeeze} and ECA-Net \cite{Wang_2020_CVPR} are using element-wise multiplication operation to recalibrate the original feature map $X$ according to the learned channel weights $Z$.

% Mapping SRM
SRM \cite{lee2019srm} adaptively squeezes the spatial dimension for the input feature map $X$ based on the style of an image via a channel-independent style pooling operator by adopting the channel-wise statistics—average and standard deviation—of each feature map as style features (i.e. d = 2). Accordingly SRM context modeling block producing style features $\textbf{Y} \in \mathbb{R}^{C\times d}$
The style features $Y_{c*d}$ are converted into channel-wise style weights $Z$ by the transformation block that consists of a style integration operator as shown in Table \ref{Tab_Equ_channel_att}. The style weights $Z$ are supposed to model the importance of the styles associated with individual channels to emphasize or suppress them accordingly.
In line with SE-Net \cite{hu2018squeeze} and ECA-Net \cite{Wang_2020_CVPR}, SRM \cite{lee2019srm} adopts the same fusion mechanism by simply employ element-wise multiplication operation to recalibrate the original feature map $X$ according to the learned channel-style weights $Z$

% Mapping GC
GCNet \cite{cao2019gcnet} adopt the same transformation block from SE-Net \cite{hu2018squeeze} to produce channel weights $Z$ while proposing a new context and fusion blocks. Given an input feature map $X$ GCNet \cite{cao2019gcnet} squeeze the channel dimension then a query-independent attention map is explicitly used for all query positions to learn the spatial relation between pixels, producing $Y$.

%%%%%%%%%%%%%%%%%%%%%%%%%%%%%%%%%%%%%%%%%%%%%%%%%%%
\begin{figure*}[t]
  \centering
  \includegraphics[width=0.95\textwidth]{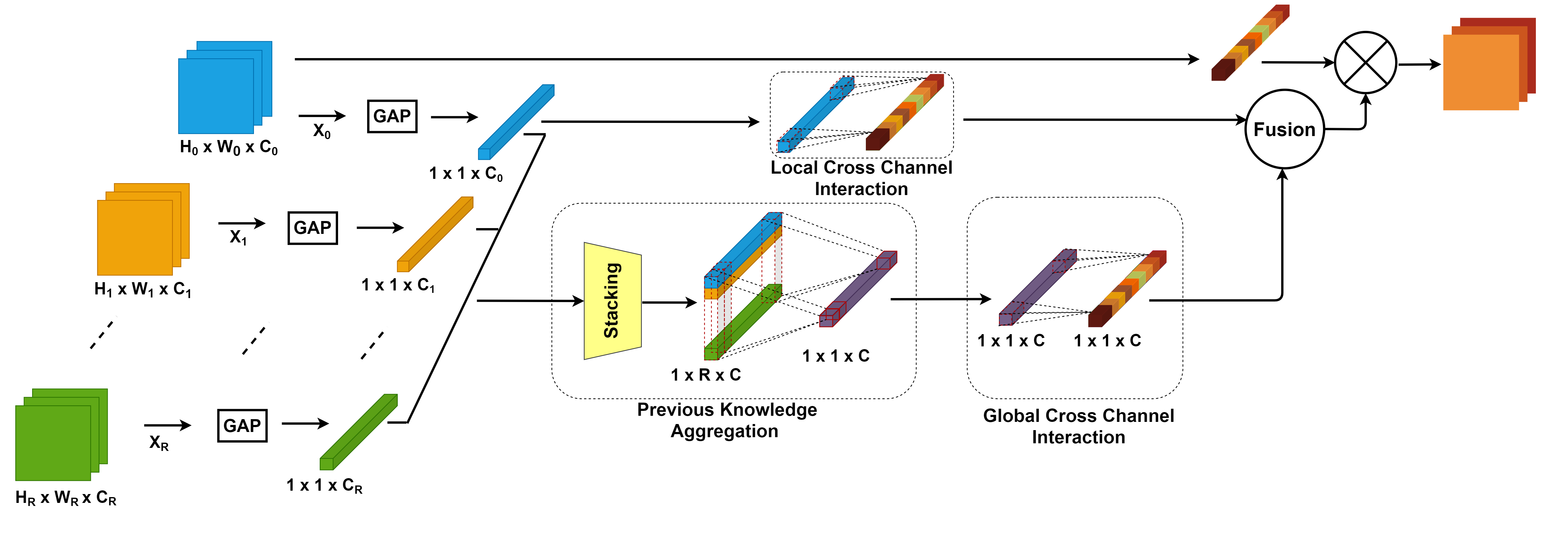}
  \caption{Diagram of our Previous Knowledge Channel Attention Module (PKCAM). Given a $R$ aggregated features, PKCAM generates global aware channel weights by performing a fast 1D convolution of size $R$, accompanied by another 1D convolution, which represents global cross channel interaction, then fused with the standard local cross channel interaction.}
  \label{Fig-MSCA_archeticture}
\end{figure*}
%%%%%%%%%%%%%%%%%%%%%%%%%%%%%%%%%%%%%%%%%%%%%%%%%%%

% -------------------------------------------------------------------------------
%                                Methodology
% -------------------------------------------------------------------------------
% 3- Previous Knowledge Channel Attention Module
\section{Methodology}
\label{Methodology}

In this section, we first demonstrate an abstracted overview of our PKCAM.
Then, we demonstrate the motivations to adopt the feature reuse concept via exploiting the previous knowledge to create a global aware attention block (i.e., PKCAM).
In addition, we develop a method to efficiently fuse both local and global cross-channel interaction modules, and finally show how to integrate it for an arbitrary CNN architecture.

\subsection{Previous Knowledge Channel Attention Module}
\label{PKCAM}

%************************************************************
% 3.2.1- {Abstracted Overview of our PKCAM}
\subsubsection{Abstracted Overview of our PKCAM}
\label{Abstracted Overview of our PKCAM}

By scrutinizing the aforementioned channel attention techniques, previous knowledge aggregation was not explored from the channel attention module perspective. Therefore we studied the previous knowledge cross-channel interaction by proposing PKCAM.

Fig.\ref{Fig-MSCA_archeticture} demonstrates our PKCAM, which exploits both local and global feature maps, through fusing two parallel attention paths while recalibrating the current feature map, where the first one consists of the local cross-channel interaction (LCCI) module that models channel-wise relationships of the aggregated feature map. We call this module local as it is operating on the current feature map only.
The second path, which we call the global path, consists of two stacked modules: previous knowledge aggregation (PKA) and global cross channel interaction (GCCI).
The PKA module covers the channel interactions across different preceding aggregated feature maps, while the GCCI module utilizes the refined features produce by the PKA module to model channel-wise relationships in a computationally efficient manner. Both LCCI and GCCI modules could be any one of the on-the-shelf channel attention techniques which are studied in Section \ref{Related work}.
%************************************************************
% 3.2.2- {Previous Knowledge Aggregation Block}
\subsubsection{Previous Knowledge Aggregation Block}
\label{Previous Knowledge Aggregation Block}

In contradiction to the aforementioned channel attention techniques which relies on the current output of an arbitrary CNN block,
our proposed PKCAM exploits both the current CNN block output
, $x_0 \in \mathbb{R}^{H_0\times W_0\times C_0}$, 
and a range of earlier CNN blocks output
, $X_p = [x_1, x_2, ..., x_R]$, 
where $R$ is the coverage region that delimits how many previous CNN blocks output will be consolidated along side the current CNN block, 
$x_1 \in \mathbb{R}^{H_1\times W_1\times C_1}$, 
$x_2 \in \mathbb{R}^{H_2\times W_2\times C_2}$, 
and $x_R \in \mathbb{R}^{H_R\times W_R\times C_R}$.

% Aligning the channel dimension among different CNN blocks
%\paragraph{Channel dimension alignment}
In general, the earlier features $X_p$ have different channel dimensions, as the conventional is as we go deeper the depth is increased.
Therefore, the first operation in our Previous Knowledge Aggregation(PKA) block is aligning the channel dimension among different CNN blocks.
As $C_0 \ge C_1 \ge C_2 \ge C_R$, aligning operation can be done be learnable upsampling techniques or a simple repeating operation to align with the channel dimension of the current CNN block $C_0$, producing channel aligned feature maps, 
$x'_0 \in \mathbb{R}^{H_0\times W_0\times C_0}$,
$x'_1 \in \mathbb{R}^{H_1\times W_1\times C_0}$, 
$x'_2 \in \mathbb{R}^{H_2\times W_2\times C_0}$, 
and $x'_R \in \mathbb{R}^{H_R\times W_R\times C_0}$.

% Aligning the spatial dimensions among different CNN blocks
%\paragraph{Spatial dimensions alignment}
Analogous to aligning the channel dimensions, the spatial dimensions; $H$ and $W$, is aligned through squeeze operation by adapting the general global average pooling equation as follows, 
$\widetilde{X} = \frac{1}{RWH}\sum_{k=1}^{R}\sum_{i=1}^{W}\sum_{j=1}^{H} X_{kij}$, 
where $\widetilde{X} \in \mathbb{R}^{R\times 1\times 1\times C_0}$
and represents the squeezed feature maps from the channel aligned aggregated feature maps $x'_i$, where $i$ = $0$, $1$, \dots, $R$, producing 
 $\widetilde{X} = [\widetilde{x}_0, \widetilde{x}_1, ..., \widetilde{x}_R]$, 
$\widetilde{x}_0 \in \mathbb{R}^{1\times 1\times C_0}$, 
$\widetilde{x}_1 \in \mathbb{R}^{1\times 1\times C_0}$ 
and $\widetilde{x}_R \in \mathbb{R}^{1\times 1\times C_0}$.

% Previous Knowledge cross-channel attention
\paragraph{Previous knowledge cross-channel attention}
Given the aggregated feature $\widetilde{X}$, previous knowledge cross-channel attention can be learned by $Y=f(\widetilde{X})$, where 
$Y \in \mathbb{R}^{1\times 1\times C_0}$
, $f(\widetilde{X}) = W' \widetilde{X}$, and $W'$ could take one of the following forms,

\begin{equation}
\label{eq_fc}
W^{'}=
\begin{cases}
   W_{1}^{'}=
    \begin{bmatrix}
        W^{'}_{1,1}                              & \cdots & W_{1,RC_0}\\
        \vdots                                   & \ddots & \vdots\\
        W^{'}_{RC_0,1} & \cdots & W^{'}_{RC_0,RC_0}
    \end{bmatrix} \\
   W_{2}^{'}=
    \begin{bmatrix}
        W^{'}_{1,1} & 0           & \cdots & 0\\
        0           & W^{'}_{2,2} &\cdots  & 0 \\
        \vdots      & \vdots      & \ddots & \vdots\\
        0           & 0           & \cdots & W^{'}_{RC_0,RC_0}\\
    \end{bmatrix}
    \end{cases}
\end{equation}
where $W'_1$ is a $RC_0 \times RC_0$ parameter matrix which learns previous knowledge interaction in conjunction with cross-channel interaction.
In contrast $W'_2$ is a $1 \times RC_0$ parameter matrix which learns previous knowledge interaction and channel interaction neglecting the cross channel relations. 
The key difference between $W'_1$ and $W'_2$ is that $W'_1$ considers previous knowledge cross-channel interaction while $W'_2$ does not, leading $W'_1$ to be more complex than $W'_2$.
Interpreting Eq. \ref{eq_fc} to neural networks $W'_1$ and $W'_2$ can be regarded as a fully connected layer and depth-wise separable convolution layer respectively. 
However, obviously from Eq. \ref{eq_fc}, $W'_1$ and $W'_2$ have a tremendous number of parameters, driving to high model complexity, especially for large channel numbers as mainly $ C_0 >> R$.

% shows the PKCAM solution
Therefore, we divide learning the previous knowledge cross-channel interaction into two sub-modules as shown in Fig. \ref{Fig-MSCA_archeticture}, learning previous knowledge interaction, and exploiting the cross-channel interaction.
Consequently, in contrast to Eq. \ref{eq_fc}, $f(\widetilde{X})$ is splitted into two cascaded functions, $f_1(\widetilde{X})$ and $f_2(\widetilde{X})$, where $f_1(\widetilde{X})$ is responsible to learn the previous knowledge channel interaction and $f_2(\widetilde{X})$ is responsible to learn the cross-channel interaction.

% Illustrate the first block --> previous knowledge channel interaction
\paragraph{Previous knowledge channel interaction}
Previous knowledge channel attention can be learned by Eq. \ref{eq_sum}, where for each channel the global information is aggregated using simple summation operation, where no learnable parameters are invoked.
\begin{equation}\label{eq_sum}
     Y = f_1(\widetilde{X}) = \sum^{C_0}_{L=1}\sum^{R}_{K=1} \widetilde{X}_{LK}
\end{equation}
A possible compromise between Eq. \ref{eq_fc} and Eq. \ref{eq_sum} is Eq. \ref{eq_ms}, where a tiny number of parameters are used whereas $W^{'} \in \mathbb{R}^{1 \times 1 \times R}$ compared to the tremendous number of parameters that are invoked in Eq. \ref{eq_fc} while learning the previous knowledge channel interaction. From the perspective of the convolution neural network, Eq. \ref{eq_ms} could be readily interpreted to a 1-D convolution layer with kernel $k = \widetilde{W}$.

\begin{equation}\label{eq_ms}
     Y = f_1(\widetilde{X}) = \sum^{C_0}_{L=1} W^{'} \widetilde{X}_{L} 
\end{equation}

% Illustrate the second block --> cross-channel interaction
\paragraph{Global cross-channel interaction}
Global cross-channel interaction could be learned by adopting one of the local channel attention modules \cite{hu2018squeeze} \cite{cao2019gcnet} \cite{wang2018non} \cite{lee2019srm} \cite{Wang_2020_CVPR} \cite{gao2019global} producing $Z_1 = f(Y)$ where $Z_1 \in \mathbb{R}^{1\times 1\times C_0}$.
\cite{Wang_2020_CVPR} \cite{gao2019global} \cite{hu2018squeeze} \cite{cao2019gcnet} \cite{wang2018non} \cite{lee2019srm} refer to the term global as they are taking into consideration the whole spatial dimension from the fed features using GAP - Global Average Pooling. In contrast we refer to the term global as previous knowledge aggregation.

%************************************************************
% 3.2.3- {Previous Knowledge Aggregation Block}
\subsubsection{Combining global and local cross-channel interaction}
\label{Combining global and local cross-channel interaction}

% Fusion mechanisms
PKCAM contains two cross-channel attention modules as shown in Figure \ref{Fig-MSCA_archeticture}, one learns the cross-channel interaction from the global feature map, while the other one learns the cross-channel interaction from the local feature map. For both modules ECA-Net \cite{Wang_2020_CVPR} is adopted empirically based on Section \ref{Ablation studies-basic local attention module}.
Finally, the current convolution output $x_0$ is recalibrated by the learned scales $S$, following Eq. \ref{eq_recalibration}, where $F \in \mathbb{R}^{1 \times 1 \times C_0}$ and $S \in \mathbb{R}^{1 \times 1 \times C_0}$ is obtained by fusing the global learned scales $Z_1$ with the local learned scales $Z_2$, following Eq. \ref{eq_fusion}.

\begin{equation}\label{eq_recalibration}
     F = \widetilde{x}_{\widetilde{W}} \odot 
\end{equation}
\begin{equation}\label{eq_fusion}
     S = \phi(Z) :Z = Concat(Z_1,Z_2)
\end{equation}

where $\phi(Z)$ represents the mapping function from the global and local scales to the final scales $S$.
The simplest mapping is summing both local and global scales as shown in Eq. \ref{eq_sum_fusion}, where no learnable parameters are included. We called this, a shallow fusion mechanism, as a fixed mapping is followed regardless of the feature map's structure and nature.
\begin{equation}
    \label{eq_sum_fusion}
    \phi(Z) = \sum_{L=1}^{C_0} \sum_{K=1}^{2} Z_{LK}
\end{equation}
Other possible mapping function is considering the full cross-channel interaction between local and global scales, which could be regarded as a fully connected layer with tremendous number of parameters.
But we waive this costly full cross-channel interaction to abide by our objective which aims to learn channel attention more efficiently in a lightweight manner besides avoiding the shallow summation fusion mechanism Eq. \ref{eq_sum_fusion}, therefore Eq. \ref{eq_ms_fusion_M_form} can be considered as a compromised mapping function, where we only concern about the channel interaction between each local channel and its counterpart in the global scales, which could be achieved easily using a 1-D convolution layer with kernel size equals two.

\begin{equation}
    \label{eq_ms_fusion_M_form}
    \phi(Z) =
    \begin{bmatrix}
    W_Z^1 & W_Z^2 \\ 
    \end{bmatrix} \otimes
    \begin{bmatrix}
    Z_1 & \cdots &  Z_1^{C_0}\\
    Z_2 & \cdots & Z_2^{C_0}
    \end{bmatrix}
\end{equation}

%************************************************************
% 3.2.4- {Integrating PKCAM Module into Deep CNNs}
\subsubsection{Integrating PKCAM Module into Deep CNNs}
\label{Integrating PKCAM Module into Deep CNNs}

%%%%%%%%%%%%%%%%%%%%%%%%%%%%%%%%%%%%%%%%%%%%%%%%%%%
\begin{figure}
  \centering
  \includegraphics[width=0.95\linewidth]{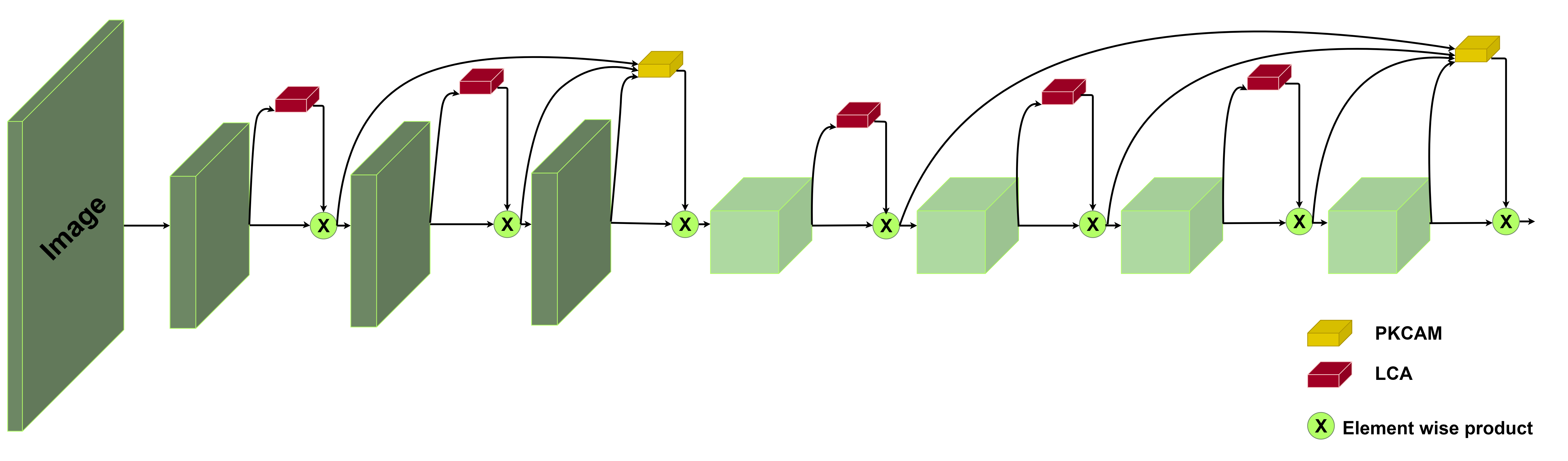}
  \caption{Illustration for the general idea of integrating PKCAM Module into an arbitrary Deep CNNs alongside local channel attention (LCA) module.}
  \label{Fig-PKCAM_integration}
\end{figure}
%%%%%%%%%%%%%%%%%%%%%%%%%%%%%%%%%%%%%%%%%%%%%%%%%%%
%\tabcolsep=0.09cm
\begin{table}
%\fontsize{8}{7.2}\selectfont
\centering
\caption{Comparison of the various ways to integrate PKCAM into CNNs using the ImageNet dataset. Top-1 accuracy is reported.}
\label{Tab-all-vs-last}
\begin{tabular}{lll}
    \toprule
Integration type &\hfil All blocks &\hfil Last block \\
    \midrule
\hfil Resnet-18 &\hfil 71.1  &\hfil \textbf{71.15} \\
\hfil Resnet-34 &\hfil 74.25  &\hfil \textbf{74.43}\\
\hfil Resnet-50 &\hfil 77.50  &\hfil \textbf{77.56} \\
\bottomrule
\end{tabular}
\end{table}
%%%%%%%%%%%%%%%%%%%%%%%%%%%%%%%%%%%%%%%%%%%%%%%%%%%
Figure \ref{Fig-PKCAM_integration} illustrate a general way for integrating PKCAM into an arbitrary CNN architecture, where PKCAM was integrated into the last CNN block for each stage alongside arbitrary local channel attention (LCA) module for the rest of the blocks. Due to the lightweight topology of our PKCAM, it could be integrated to each block, where LCA is totally replaced.
%Both approaches have shown comparable results as shown in Table 
%%%%%%%%%%%%%%%%%%%%%%%%%%%%%%%%%%%%%%%%%%%%%%%%%%%
\begin{figure}
  \centering
  \includegraphics[width=0.95\linewidth, frame]{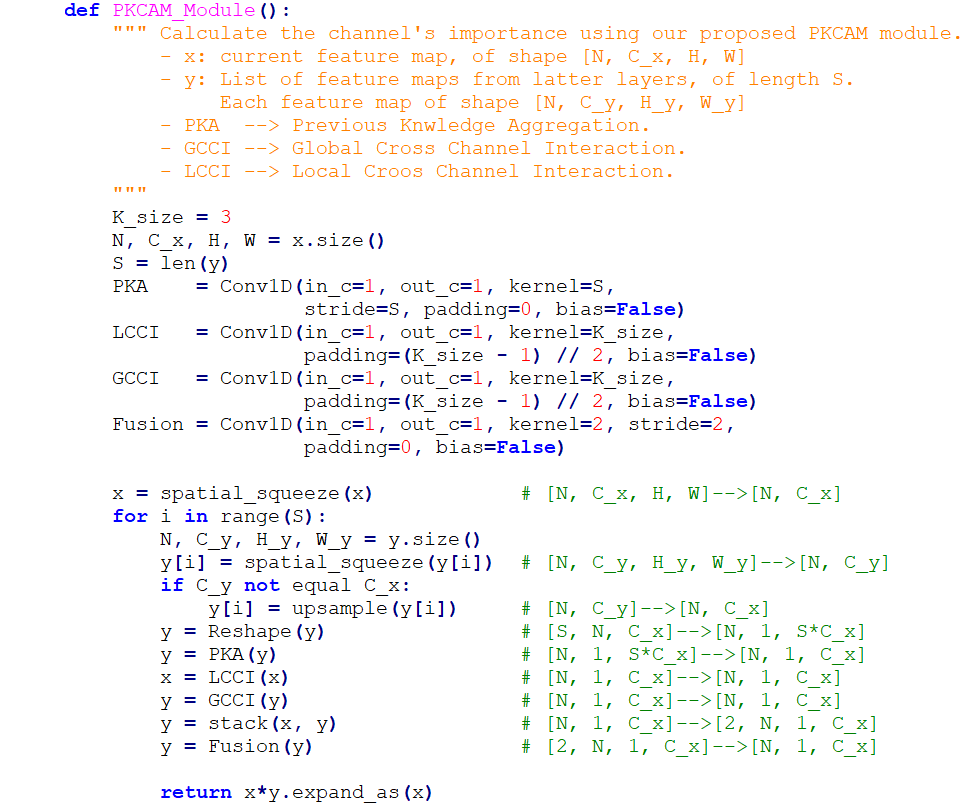}
  \caption{Pseudo code for our PKCAM.}
  \label{Fig-PKCAM_Pseudo_code}
\end{figure}
%%%%%%%%%%%%%%%%%%%%%%%%%%%%%%%%%%%%%%%%%%%%%%%%%%%
Figure \ref{Fig-PKCAM_Pseudo_code} demonstrate a pseudo code for our PKCAM to show how easily it could be integrated to any CNN architecture.

% -------------------------------------------------------------------------------
%                                Experiments
% -------------------------------------------------------------------------------
%%%%%%%%%%%%%%%%%%%%%%%%%%%%%%%%%%%%%%%%%%%%
% Table start
%\tabcolsep=0.08cm
\begin{table*}[t]
  \caption{Comparison of different previous knowledge Aggregation(PKA) techniques on the horizontal dimension, where 1-D Conv., Sum and FC stands for one dimensional convolution layer Eq.\ref{eq_ms}, summation Eq.\ref{eq_sum}, and fully connected layer Eq.\ref{eq_fc} respectively. While the vertical dimension compares various global-local fusion mechanisms, where 1-D Conv., Sum and FC stands for one dimensional convolution layer Eq.\ref{eq_ms_fusion_M_form}, summation Eq.\ref{eq_sum_fusion}, and fully connected layer respectively.}
  \label{Table-Ablation studies-Multi-scale}
  \centering
  \scalebox{0.9}{
  \begin{tabular}{llllllllllllll}
    \toprule
 \multicolumn{2}{c}{\backslashbox{Fusion}{MSCI}} &\-  & \hfil \makecell[c]{1-D \\ Conv.} & \hfil Sum &\hfil FC & \- &\hfil \makecell[c]{1-D \\ Conv.} & \hfil Sum &\hfil FC & \- &\hfil \makecell[c]{1-D \\ Conv.}  & \hfil Sum &\hfil FC\\
    %\midrule
    \cmidrule(r){1-3}
    \cmidrule(r){4-6}
    \cmidrule(r){8-10}
    \cmidrule(r){12-14}
    \- & \- & \- & \multicolumn{3}{c}{ResNet-18} & \- & \multicolumn{3}{c}{ResNet-34} & \- & \multicolumn{3}{c}{ResNet-50} \\
    \cmidrule(r){4-6}
    \cmidrule(r){8-10}
    \cmidrule(r){12-14}
     
    \- &\hfil Acc. &\- &\hfil 55.70 &\hfil 55.28 &\hfil  54.63 & \- &\hfil  \textbf{56.94}   & \hfil 56.26  & \hfil 56.52  & \-  & \hfil \textbf{57.89} &\hfil  56.18 & \hfil 56.41 \\  
      \cmidrule(r){2-3}
    \hfil \makecell[c]{1-D \\ Conv.} &\hfil \#.P (M) &\- &\hfil \textbf{10.749} &\hfil 10.749   &\hfil 11.413 & \- & \hfil \textbf{20.389}   & \hfil 20.389  & \hfil 23.049  & \-  & \hfil \textbf{22.824} &\hfil 22.824 &\hfil 65.387 \\ 
      \cmidrule(r){2-3}
    \-&\hfil GFLOPs &\- &\hfil \textbf{2.075} &\hfil 2.075 &\hfil 2.076 & \- &\hfil \textbf{4.329}  &\hfil 4.329 &\hfil 4.331  & \-  & \hfil \textbf{4.878} &\hfil 4.878 &\hfil 4.878  \\  

     \cmidrule(r){1-14}
        
    \- &\hfil Acc. &\- & \hfil 54.96 &\hfil  55.10 & \hfil 54.24 & \- &\hfil 56.00   &\hfil 56.02  &\hfil 55.40  & \-& \hfil 57.14 &\hfil 56.55 & \hfil 57.16? \\  
    \cmidrule(r){2-3}
   \hfil Sum &\hfil \#.P (M) &\- & \hfil 10.749 & \hfil 10.749   & \hfil 11.413 & \- & \hfil 20.389   & \hfil 20.389  &\hfil 23.049  & \-  & \hfil 22.824 &\hfil 22.824 &\hfil 65.387 \\ 
      \cmidrule(r){2-3}
    \-&\hfil GFLOPs &\- & \hfil 2.075 & \hfil 2.075 &\hfil  2.076 & \- & \hfil 4.329 &\hfil 4.329 & \hfil 4.331 & \-  &\hfil  4.878 &\hfil 4.878 &\hfil 4.878  \\  

      \cmidrule(r){1-14}
        
    \- &\hfil Acc. &\- & \hfil \textbf{56.01} &\hfil 55.35 &\hfil  54.41 & \- & \hfil 56.20   & \hfil 56.38  & \hfil 56.64  & \- &\hfil 57.53 &\hfil  56.69 &  \hfil 56.95 \\   
    \cmidrule(r){2-3}
    \hfil FC &\hfil \#.P (M) &\- &\hfil 11.413 &\hfil 11.413 &\hfil 12.077 & \- &\hfil 22.124   &\hfil 22.124  &\hfil 24.784  & \-  & \hfil 50.574 &\hfil 50.574 &\hfil 93.137 \\
      \cmidrule(r){2-3}
    \-&\hfil GFLOPs &\- &\hfil 2.076 &\hfil 2.076 & \hfil 2.076 & \- &\hfil 4.330   &\hfil 4.330  & \hfil 4.333  & \-  &\hfil 4.906 &\hfil 4.906 &\hfil 4.947  \\  

    \bottomrule
  \end{tabular}
  }
\end{table*}
% Table End
%%%%%%%%%%%%%%%%%%%%%%%%%%%%%%%%%%%%%%%%%%%%
\section{Experiments}
\label{Experiments}
In this section, we perform controlled ablation experiments to settle on the best design for our proposed module and assess its sub-modules.
Then we evaluate the performance of the proposed Previous Knowledge Attention module, on a series of benchmark datasets across different tasks including Tiny-ImageNet \cite{le2015tiny}, and ImageNet \cite{deng2009imagenet} for the classification task, and KITTI \cite{Geiger2012CVPR} for detection.
%and CityScapes \cite{cordts2016cityscapes} for segmentation task.
Finally, We conduct empirical experiments that probe the robustness of the representations learned by PKCAM, compared to convolutional baselines and other attention mechanisms.

% -----------------------------------------------------------
% 1- Ablation studies and analysis
\subsection{Ablation studies and analysis}
\label{Ablation studies}
We have conducted three ablations studies to settle on the best architecture design and analyze the effectiveness of each component in our PKCAM. The first one investigating the different approaches for previous knowledge channel interaction that are discussed in Section \ref{Previous Knowledge Aggregation Block}.
Then, we assess the choice of basic attention modules that are used in the local and global cross channel interaction modules as shown in Fig. \ref{Fig-MSCA_archeticture}.
Finally, we demonstrate the effectiveness of our proposed previous knowledge cross-channel interaction module compared to the naive local cross-channel interaction.
%Then, we revisit the effect of the kernel size (k) in ECA Module \cite{Wang_2020_CVPR}, and show that our proposed method achieves consistent improvements in performances.

% -----------------------------------------------------------
% 1.1- Global and local cross-channel interaction
\subsubsection{Global and local cross-channel interaction}
\label{Ablation studies-global-local}

We have investigated empirically the different global channel interaction techniques that were described in detail in Section\ref{Previous Knowledge Aggregation Block}, besides examined the various global-local fusion mechanisms that combine the global and the local cross-channel learned representations which were cover in Section\ref{Combining global and local cross-channel interaction}.
To cover the whole possible combinations of each sub-module, nine experiments are conducted for each adopted backbone, which are ResNet-18 \cite{he2016deep}, ResNet-34 \cite{he2016deep}, ResNet-50 \cite{he2016deep}.
All experiments in Table \ref{Table-Ablation studies-Multi-scale} are conducted using the Tiny-ImageNet dataset \cite{le2015tiny}, where the same data augmentation and hyper-parameter settings in \cite{hu2018squeeze} are adopted. Input images are randomly cropped to $64\times64$ with random horizontal flipping. Stochastic gradient descent (SGD) with weight decay of $1e{-4}$, the momentum of 0.9, and mini-batch size of 32 is used.
Models are trained for 100 epochs from scratch, using the weight initialization strategy described in \cite{he2015delving} and the initial learning rate is set to 0.1 and decreased by a factor of 10 every 30 epochs.
As shown in Table\ref{Table-Ablation studies-Multi-scale} exploiting the previous knowledge channel interaction through 1-D Conv. layer by following Eq.\ref{eq_ms}, and fuse the global and local scales by following Eq.\ref{eq_ms_fusion_M_form}, is the best compromise to solve the paradox of performance and complexity trade-off, where it shares almost the same model complexity (i.e., network parameters and FLOPs) with the original ResNet while at the same time it achieves the best accuracy for ResNet-50 and comparable accuracy for ResNet-18 and ResNet-34 compared to fusing the global and local scales using fully connected layer which invokes a tremendous number of parameters.
Based on the aforementioned results in Table \ref{Table-Ablation studies-Multi-scale}, our novel approach follows the compromised combination while exploiting the previous knowledge cross channel interaction by following Eq.\ref{eq_ms} to capture the previous knowledge channel interaction and Eq.\ref{eq_ms_fusion_M_form} to fuse the global and local learned representations.

% -----------------------------------------------------------
%\tabcolsep=0.09cm
\begin{table}
%\fontsize{8}{7.2}\selectfont
\centering
\caption{Comparison of the various basic attention modules in our proposed module PKCAM using the Tiny-ImageNet dataset.}
\label{Tab-basic-attention-module}
\begin{tabular}{llll}
    \toprule
Basic Attention Module &\hfil SE &\hfil SRM \hfil& ECA\\
    \midrule
\hfil Resnet-18 &\hfil 54.07 &\hfil 55.10 &\hfil \textbf{55.70} \\
\hfil Resnet-34 &\hfil 56.61 &\hfil 56.52 &\hfil \textbf{56.94}\\
\hfil Resnet-50 &\hfil 57.42     &\hfil 57.12 &\hfil \textbf{57.89} \\
\bottomrule
\end{tabular}
\end{table}

% -----------------------------------------------------------
% 1.2- basic local attention module
\subsubsection{Basic attention module}
\label{Ablation studies-basic local attention module}

We next assess the choice of basic attention modules that are used in the local and global cross channel interaction modules as shown in Fig.\ref{Fig-MSCA_archeticture}.
Three channel attention mechanisms are evaluated on the Tiny-
Imagenet dataset \cite{le2015tiny}, including SE-Net \cite{hu2018squeeze}, SRM \cite{lee2019srm}, and ECA-Net \cite{Wang_2020_CVPR}. As shown in Table \ref{Tab-basic-attention-module} ECA-Net achieves the best accuracy across different ResNet backbones. However, building up our PKCAM using other channel attention mechanism boost the accuracy compared to their original results. 
For example, SRM using ResNet-18 as a backbone achieves 53.39\% while our PKCAM module builds upon SRM achieves 55.1\%.

\tabcolsep=0.03cm
\begin{table*}
\fontsize{8}{7.2}\selectfont
  \caption{Comparisons with state-of-the-art attention modules on ImageNet in terms of the number of parameters (\#P.) in millions, GFLOPs, top-1, and top-5 accuracy. Top-1 relative improvement results is reported between parentheses w.r.t SENet improvement over Vanilla Resnet.}
  \label{Table-imagenet results}
  \centering
  \begin{tabular}{llllllllllllllll}
    \toprule
    Methods      & \- &\hfil \#.P.(M) &\hfil GFLOPs &\hfil Top-1 &\hfil Top-5  &\- &\hfil\#.P.(M) &\hfil GFLOPs &\hfil Top-1 &\hfil Top-5 &\- &\hfil\#.P.(M) &\hfil GFLOPs &\hfil Top-1 &\hfil Top-5 \\
     %\midrule
      \cmidrule(r){1-2}
      \cmidrule(r){3-6}
      \cmidrule(r){7-11}
      \cmidrule(r){12-16}
    \-&\-&\multicolumn{4}{c}{ResNet-18}&\-&\multicolumn{4}{c}{ResNet-34} &\-&\multicolumn{4}{c}{ResNet-50}\\
    %\multicolumn{1}{c}{\-} &  \- & \- & ResNet-18  &  \- &  \- & \-&\- &  ResNet-34 & \- & \- \\
    \cmidrule(r){3-6}
    \cmidrule(r){7-11}
    \cmidrule(r){12-16}
    ResNet \cite{he2016deep}   & \- & \hfil 11.14   & \hfil 1.699  &\hfil 70.42    &\hfil 89.45   & \-  &\hfil 20.78         &\hfil 3.427  &\hfil 73.31   &\hfil 91.40 & \-  &\hfil 24.37         &\hfil 3.86  &\hfil 75.2   &\hfil 92.52   \\  
    SENet \cite{hu2018squeeze}    & \- & \hfil 11.23   & \hfil 1.700  &\hfil 70.59    &\hfil 89.78   & \-  &\hfil 20.93         &\hfil 3.428  &\hfil 73.87   &\hfil 91.65 & \-  &\hfil 26.77         &\hfil 3.87  &\hfil 76.71   &\hfil 93.38\\
    CBAM \cite{woo2018cbam}     & \- & \hfil 11.23   & \hfil 1.700  &\hfil 70.73(182\%)    &\hfil 89.91   & \-  &\hfil 20.94         &\hfil 3.428  &\hfil 74.01(125\%)   &\hfil 91.76 & \-  &\hfil 26.77         &\hfil 3.87  &\hfil 77.34(141\%)   &\hfil 93.69  \\
    ECA \cite{Wang_2020_CVPR}   & \- & \hfil 11.14   & \hfil 1.699  &\hfil 70.78(211\%)    &\hfil 89.92   & \-  &\hfil 20.78         &\hfil 3.427  &\hfil 74.21(160\%)   &\hfil 91.83 & \-  &\hfil 24.37         &\hfil 3.86  &\hfil 77.48(151\%)   &\hfil 93.68  \\
    PKCAM    & \- & \hfil \textbf{11.14}   & \hfil \textbf{1.699}  &\hfil \textbf{70.98(329\%)}    &\hfil \textbf{90.12}       & \-  &\hfil \textbf{20.78}    &\hfil \textbf{3.427}  &\hfil \textbf{74.43(200\%)}       &\hfil \textbf{91.87} & \-  &\hfil \textbf{24.37}         &\hfil \textbf{3.86}  &\hfil \textbf{77.56(156\%)}   &\hfil \textbf{93.70}      \\

   % \midrule
  
    \bottomrule
  \end{tabular}
\end{table*}
%%%%%%%%%%%%%%%%%%%%%%%%%%%%%%%%%%%%%%%%%%%%%%%%%%%
% -----------------------------------------------------------
% 1.3- Effect of local vs. global
\subsubsection{Local Vs. global cross channel interaction}
\label{Ablation studies-local vs. global}

We conduct experiments to validate the effectiveness of our proposed global cross-channel interaction module by comparing it with the naive local cross-channel interaction.
Experiments at Table \ref{Tab-local-global} are conducted using the Tiny-Imagenet dataset \cite{le2015tiny} and ECA-Net \cite{Wang_2020_CVPR} as basic channel module as discussed at Section \ref{Ablation studies-basic local attention module}.
Table \ref{Tab-local-global} shows that the global cross-channel interaction module alone achieves better accuracy than the local one, and combining both of them achieves the best accuracy.
Results that are shown in Table \ref{Tab-local-global} suggest that the recalibration scales learning process will benefit from global information.

%************************************************************
% 2- Integrating MSCA Module into Deep CNNs
% \subsection{Integrating MSCA Module into Deep CNNs [Future Work]}
% \label{Integrating MSCA Module into Deep CNNs}

% ---------------------------------------------------------------------------

%\tabcolsep=0.08cm
\begin{table}
\begin{minipage}[c]{0.5\textwidth}
%\fontsize{8}{7.2}\selectfont
\caption{Showing effectiveness of previous knowledge cross-channel interaction module using the Tiny-ImageNet dataset.}
\label{Tab-local-global}
\scalebox{0.9}{
\begin{tabular}{llll}
    \toprule
Local Vs. Global &\hfil Local &\hfil Global &\hfil Both\\
    \midrule
\hfil Resnet-18 &\hfil 53.76 &\hfil 54.82    &\hfil \textbf{55.70} \\
\hfil Resnet-34 &\hfil 55.66 &\hfil 56.18    &\hfil \textbf{56.94} \\
\hfil Resnet-50 &\hfil 56.59 &\hfil 56.89        &\hfil \textbf{57.89}\\
\bottomrule
\end{tabular}
}
\end{minipage}
\begin{minipage}[c]{0.5\textwidth}
\caption{Comparisons with state-of-the-art attention modules on KITTI-RGB in terms of mAP using YOLOV3 on Resnet-18 and 34 backbones.}
\label{Table_objectdetection_results}
  \scalebox{0.8}{
  \centering
  \begin{tabular}{llllllll}
    \toprule
     \hfil  &\hfil Vanilla &\hfil SE &\hfil ECA &\hfil CBAM & \hfil BAM &\hfil SRM &\hfil PKCAM \\
    \midrule
   \hfil R-18 &\hfil 57.87 &\hfil 59.32 &\hfil 58.55 &\hfil 57.90 &\hfil 59.61 &\hfil 59.20 &\hfil \textbf{59.66} \\
   \hfil R-50 &\hfil 64.19 &\hfil 65.08 &\hfil 64.34 &\hfil 64.18 &\hfil 65.10 &\hfil 64.82 &\hfil \textbf{65.21}\\
    \bottomrule
  \end{tabular}
  }
\end{minipage}
\end{table}
%%%%%%%%%%%%%%%%%%%%%%%%%%%%%%%%%%%%%%%%%
% ImageNet
\subsection{Image classification on ImageNet}
\label{Image classification on ImageNet}

In this section, we evaluate the performance of proposed PKCAM network on ImageNet \cite{deng2009imagenet}.
All the classification experiments follows the same training procedure, where the same data augmentation and hyper-parameter settings in \cite{hu2018squeeze} are adopted.
Models are trained for 100 epochs from scratch, using the weight initialization strategy described in \cite{he2015delving} and the initial learning rate is set to 0.1 and decreased by a factor of 10 every 30 epochs.
Stochastic gradient descent (SGD) with weight decay of $1e{-4}$, the momentum of 0.9, and mini-batch size of 256 for ImageNet \cite{deng2009imagenet}.
The evaluation metrics incorporate both efficiencies (i.e., network parameters (\#P.) in millions, and floating-point operations per second (FLOPs) in Gigas) and effectiveness (i.e., Top-1 accuracy).

ImageNet LSVRC 2012 dataset \cite{deng2009imagenet}, which contains $10^{3}$ classes with 1.2 million training images, $50\times10^{3}$ validation images, and $100\times10^{3}$ test images. The evaluation is measured on the non-blacklist images of the ImageNet LSVRC 2012 validation set. 
A $224\times224$ crop is randomly sampled from an image or its horizontal flip, with the per-pixel RGB mean value subtracted.

We compare our PKCAM module with several state-of-the-art attention methods using ResNet-18 and ResNet-34 backbones \cite{he2016deep} on ImageNet.
Efficiency and effectiveness are measured, and the results are reported in Table \ref{Table-imagenet results} from their original papers. We adopt the same training setup as \cite{he2016deep} \cite{hu2018squeeze} for fair comparison.
Results show that our proposed PKCAM achieves the best accuracy besides be the lightest model compared to other attention modules \cite{hu2018squeeze} \cite{woo2018cbam}.
Top-1 relative improvement results is reported between parentheses w.r.t SENet improvement over Vanilla Resnet.

% -----------------------------------------------------------
% 4- Object Detection
\subsection{Object detection}

KITTI-RGB \cite{Geiger2012CVPR} consists of 7,481 training images and 7,518 test images, comprising a total of 80,256 labeled objects of eight different classes. Each image has 3 RGB color channels and pixel dimensions $1242\times375$ which is resized to $224\times224$.
We follow the same training setup as mentioned at Section \ref{Image classification on ImageNet}.
As shown in Table \ref{Table_objectdetection_results}, PKCAM considerably improves the accuracy more than other attention modules compared to the baseline \cite{he2016deep}. YOLOV3 \cite{redmon2018yolov3} detector is used.

% -----------------------------------------------------------
% 5- Semantic Segmentation
% \subsection{Semantic segmentation [Future Work]}

% -----------------------------------------------------------
% 6- Model And Computational Complexity
%\subsection{Model and computational complexity [Future Work]}
%\label{computational complexity}

% -----------------------------------------------------------
% 6- Pruning
% \subsection{Pruning}
% \label{Pruning}

% -------------------------------------------------------------------------------
%                                Conclusion
% -------------------------------------------------------------------------------
\section{Conclusion}
\label{Conclusion}
In this paper, we concentrate on determining an effective channel attention module with low model complexity. To this end, we propose efficient channel attention (PKCAM).
Because of the lightweight computation of the PKCAM block, it can be integrated into all modern CNN architectures across the whole layers and trained end-to-end.
While most previous works utilized uni-scale features, PKCAM is designed to employ the ability of global information while recalibrating feature maps.
Our experiments demonstrate that simply inserting PKCAM into standard CNN architectures boosts the performance across different tasks.
Furthermore, we verify the robustness of the representations learned by PKCAM and its generalization ability via zero-shot experiments to rotated images.
% -------------------------------------------------------------------------------

% -------------------------------------------------------------------------------
{\small
\bibliographystyle{ieee_fullname}
\bibliography{egbib}

\begin{thebibliography}{10}\itemsep=-1pt

\bibitem{beck2009top}
Diane~M Beck and Sabine Kastner.
\newblock Top-down and bottom-up mechanisms in biasing competition in the human
  brain.
\newblock {\em Vision research}, 49(10):1154--1165, 2009.

\bibitem{cao2019gcnet}
Yue Cao, Jiarui Xu, Stephen Lin, Fangyun Wei, and Han Hu.
\newblock Gcnet: Non-local networks meet squeeze-excitation networks and
  beyond.
\newblock In {\em Proceedings of the IEEE/CVF International Conference on
  Computer Vision Workshops}, pages 0--0, 2019.

\bibitem{carion2020end}
Nicolas Carion, Francisco Massa, Gabriel Synnaeve, Nicolas Usunier, Alexander
  Kirillov, and Sergey Zagoruyko.
\newblock End-to-end object detection with transformers.
\newblock In {\em European Conference on Computer Vision}, pages 213--229.
  Springer, 2020.

\bibitem{chen2017sca}
Long Chen, Hanwang Zhang, Jun Xiao, Liqiang Nie, Jian Shao, Wei Liu, and
  Tat-Seng Chua.
\newblock Sca-cnn: Spatial and channel-wise attention in convolutional networks
  for image captioning.
\newblock In {\em Proceedings of the IEEE conference on computer vision and
  pattern recognition}, pages 5659--5667, 2017.

\bibitem{chen2016attention}
Liang-Chieh Chen, Yi Yang, Jiang Wang, Wei Xu, and Alan~L Yuille.
\newblock Attention to scale: Scale-aware semantic image segmentation.
\newblock In {\em Proceedings of the IEEE conference on computer vision and
  pattern recognition}, pages 3640--3649, 2016.

\bibitem{deng2009imagenet}
Jia Deng, Wei Dong, Richard Socher, Li-Jia Li, Kai Li, and Li Fei-Fei.
\newblock Imagenet: A large-scale hierarchical image database.
\newblock In {\em 2009 IEEE conference on computer vision and pattern
  recognition}, pages 248--255. Ieee, 2009.

\bibitem{desimone1998visual}
Robert Desimone.
\newblock Visual attention mediated by biased competition in extrastriate
  visual cortex.
\newblock {\em Philosophical Transactions of the Royal Society of London.
  Series B: Biological Sciences}, 353(1373):1245--1255, 1998.

\bibitem{desimone1995neural}
Robert Desimone and John Duncan.
\newblock Neural mechanisms of selective visual attention.
\newblock {\em Annual review of neuroscience}, 18(1):193--222, 1995.

\bibitem{fu2019dual}
Jun Fu, Jing Liu, Haijie Tian, Yong Li, Yongjun Bao, Zhiwei Fang, and Hanqing
  Lu.
\newblock Dual attention network for scene segmentation.
\newblock In {\em Proceedings of the IEEE/CVF Conference on Computer Vision and
  Pattern Recognition}, pages 3146--3154, 2019.

\bibitem{gao2019global}
Zilin Gao, Jiangtao Xie, Qilong Wang, and Peihua Li.
\newblock Global second-order pooling convolutional networks.
\newblock In {\em Proceedings of the IEEE/CVF Conference on Computer Vision and
  Pattern Recognition}, pages 3024--3033, 2019.

\bibitem{Geiger2012CVPR}
Andreas Geiger, Philip Lenz, and Raquel Urtasun.
\newblock Are we ready for autonomous driving? the kitti vision benchmark
  suite.
\newblock In {\em Conference on Computer Vision and Pattern Recognition
  (CVPR)}, 2012.

\bibitem{he2015delving}
Kaiming He, Xiangyu Zhang, Shaoqing Ren, and Jian Sun.
\newblock Delving deep into rectifiers: Surpassing human-level performance on
  imagenet classification.
\newblock In {\em Proceedings of the IEEE international conference on computer
  vision}, pages 1026--1034, 2015.

\bibitem{he2016deep}
Kaiming He, Xiangyu Zhang, Shaoqing Ren, and Jian Sun.
\newblock Deep residual learning for image recognition.
\newblock In {\em Proceedings of the IEEE conference on computer vision and
  pattern recognition}, pages 770--778, 2016.

\bibitem{hu2018gather}
Jie Hu, Li Shen, Samuel Albanie, Gang Sun, and Andrea Vedaldi.
\newblock Gather-excite: Exploiting feature context in convolutional neural
  networks.
\newblock {\em arXiv preprint arXiv:1810.12348}, 2018.

\bibitem{hu2018squeeze}
Jie Hu, Li Shen, and Gang Sun.
\newblock Squeeze-and-excitation networks.
\newblock In {\em Proceedings of the IEEE conference on computer vision and
  pattern recognition}, pages 7132--7141, 2018.

\bibitem{huang2017densely}
Gao Huang, Zhuang Liu, Laurens Van Der~Maaten, and Kilian~Q Weinberger.
\newblock Densely connected convolutional networks.
\newblock In {\em Proceedings of the IEEE conference on computer vision and
  pattern recognition}, pages 4700--4708, 2017.

\bibitem{iandola2014densenet}
Forrest Iandola, Matt Moskewicz, Sergey Karayev, Ross Girshick, Trevor Darrell,
  and Kurt Keutzer.
\newblock Densenet: Implementing efficient convnet descriptor pyramids.
\newblock {\em arXiv preprint arXiv:1404.1869}, 2014.

\bibitem{krizhevsky2012imagenet}
Alex Krizhevsky, Ilya Sutskever, and Geoffrey~E Hinton.
\newblock Imagenet classification with deep convolutional neural networks.
\newblock {\em Advances in neural information processing systems},
  25:1097--1105, 2012.

\bibitem{le2015tiny}
Ya Le and Xuan Yang.
\newblock Tiny imagenet visual recognition challenge.
\newblock {\em CS 231N}, 7, 2015.

\bibitem{lee2019srm}
HyunJae Lee, Hyo-Eun Kim, and Hyeonseob Nam.
\newblock Srm: A style-based recalibration module for convolutional neural
  networks.
\newblock In {\em Proceedings of the IEEE/CVF International Conference on
  Computer Vision}, pages 1854--1862, 2019.

\bibitem{linsley2018learning}
Drew Linsley, Dan Shiebler, Sven Eberhardt, and Thomas Serre.
\newblock Learning what and where to attend.
\newblock {\em arXiv preprint arXiv:1805.08819}, 2018.

\bibitem{luo2016understanding}
Wenjie Luo, Yujia Li, Raquel Urtasun, and Richard Zemel.
\newblock Understanding the effective receptive field in deep convolutional
  neural networks.
\newblock In {\em Proceedings of the 30th International Conference on Neural
  Information Processing Systems}, pages 4905--4913, 2016.

\bibitem{ma2015hierarchical}
Chao Ma, Jia-Bin Huang, Xiaokang Yang, and Ming-Hsuan Yang.
\newblock Hierarchical convolutional features for visual tracking.
\newblock In {\em Proceedings of the IEEE international conference on computer
  vision}, pages 3074--3082, 2015.

\bibitem{mnih2014recurrent}
Volodymyr Mnih, Nicolas Heess, Alex Graves, and Koray Kavukcuoglu.
\newblock Recurrent models of visual attention.
\newblock {\em arXiv preprint arXiv:1406.6247}, 2014.

\bibitem{newell2016stacked}
Alejandro Newell, Kaiyu Yang, and Jia Deng.
\newblock Stacked hourglass networks for human pose estimation.
\newblock In {\em European conference on computer vision}, pages 483--499.
  Springer, 2016.

\bibitem{park2018bam}
Jongchan Park, Sanghyun Woo, Joon-Young Lee, and In~So Kweon.
\newblock Bam: Bottleneck attention module.
\newblock {\em arXiv preprint arXiv:1807.06514}, 2018.

\bibitem{redmon2018yolov3}
Joseph Redmon and Ali Farhadi.
\newblock Yolov3: An incremental improvement.
\newblock {\em arXiv preprint arXiv:1804.02767}, 2018.

\bibitem{ronneberger2015u}
Olaf Ronneberger, Philipp Fischer, and Thomas Brox.
\newblock U-net: Convolutional networks for biomedical image segmentation.
\newblock In {\em International Conference on Medical image computing and
  computer-assisted intervention}, pages 234--241. Springer, 2015.

\bibitem{roy2018recalibrating}
Abhijit~Guha Roy, Nassir Navab, and Christian Wachinger.
\newblock Recalibrating fully convolutional networks with spatial and channel
  “squeeze and excitation” blocks.
\newblock {\em IEEE transactions on medical imaging}, 38(2):540--549, 2018.

\bibitem{simonyan2014very}
Karen Simonyan and Andrew Zisserman.
\newblock Very deep convolutional networks for large-scale image recognition.
\newblock {\em arXiv preprint arXiv:1409.1556}, 2014.

\bibitem{srivastava2015training}
Rupesh~Kumar Srivastava, Klaus Greff, and J{\"u}rgen Schmidhuber.
\newblock Training very deep networks.
\newblock {\em arXiv preprint arXiv:1507.06228}, 2015.

\bibitem{szegedy2017inception}
Christian Szegedy, Sergey Ioffe, Vincent Vanhoucke, and Alexander Alemi.
\newblock Inception-v4, inception-resnet and the impact of residual connections
  on learning.
\newblock In {\em Proceedings of the AAAI Conference on Artificial
  Intelligence}, volume~31, 2017.

\bibitem{szegedy2015going}
Christian Szegedy, Wei Liu, Yangqing Jia, Pierre Sermanet, Scott Reed, Dragomir
  Anguelov, Dumitru Erhan, Vincent Vanhoucke, and Andrew Rabinovich.
\newblock Going deeper with convolutions.
\newblock In {\em Proceedings of the IEEE conference on computer vision and
  pattern recognition}, pages 1--9, 2015.

\bibitem{szegedy2016rethinking}
Christian Szegedy, Vincent Vanhoucke, Sergey Ioffe, Jon Shlens, and Zbigniew
  Wojna.
\newblock Rethinking the inception architecture for computer vision.
\newblock In {\em Proceedings of the IEEE conference on computer vision and
  pattern recognition}, pages 2818--2826, 2016.

\bibitem{vaswani2017attention}
Ashish Vaswani, Noam Shazeer, Niki Parmar, Jakob Uszkoreit, Llion Jones,
  Aidan~N Gomez, {\L}ukasz Kaiser, and Illia Polosukhin.
\newblock Attention is all you need.
\newblock In {\em Advances in neural information processing systems}, pages
  5998--6008, 2017.

\bibitem{wang2017residual}
Fei Wang, Mengqing Jiang, Chen Qian, Shuo Yang, Cheng Li, Honggang Zhang,
  Xiaogang Wang, and Xiaoou Tang.
\newblock Residual attention network for image classification.
\newblock In {\em Proceedings of the IEEE conference on computer vision and
  pattern recognition}, pages 3156--3164, 2017.

\bibitem{Wang_2020_CVPR}
Qilong Wang, Banggu Wu, Pengfei Zhu, Peihua Li, Wangmeng Zuo, and Qinghua Hu.
\newblock Eca-net: Efficient channel attention for deep convolutional neural
  networks.
\newblock In {\em Proceedings of the IEEE/CVF Conference on Computer Vision and
  Pattern Recognition (CVPR)}, June 2020.

\bibitem{wang2018non}
Xiaolong Wang, Ross Girshick, Abhinav Gupta, and Kaiming He.
\newblock Non-local neural networks.
\newblock In {\em Proceedings of the IEEE conference on computer vision and
  pattern recognition}, pages 7794--7803, 2018.

\bibitem{woo2018cbam}
Sanghyun Woo, Jongchan Park, Joon-Young Lee, and In~So Kweon.
\newblock Cbam: Convolutional block attention module.
\newblock In {\em Proceedings of the European conference on computer vision
  (ECCV)}, pages 3--19, 2018.

\bibitem{zhao2020exploring}
Hengshuang Zhao, Jiaya Jia, and Vladlen Koltun.
\newblock Exploring self-attention for image recognition.
\newblock In {\em Proceedings of the IEEE/CVF Conference on Computer Vision and
  Pattern Recognition}, pages 10076--10085, 2020.

\end{thebibliography}
}

\end{document}